\documentclass[10pt,twocolumn,letterpaper]{article}
\usepackage{cvpr}

\usepackage{graphicx}
\usepackage{amsmath}
\usepackage{amssymb}
\usepackage{booktabs}
\usepackage{multirow}
\usepackage{pifont}

\usepackage{algorithm}
\usepackage{algorithmic}

\usepackage[table]{xcolor}
\definecolor{mygray}{gray}{.92}
\definecolor{mycyan}{cmyk}{.3,0,0,0}
\definecolor{LightCyan}{rgb}{0.88,1,1}

\makeatletter
\newcommand{\thickhline}{%
    \noalign {\ifnum 0=`}\fi \hrule height 1pt
    \futurelet \reserved@a \@xhline}

\usepackage[pagebackref,breaklinks,colorlinks]{hyperref}

\usepackage[capitalize]{cleveref}
\crefname{section}{Sec.}{Secs.}
\Crefname{section}{Section}{Sections}
\Crefname{table}{Table}{Tables}
\crefname{table}{Tab.}{Tabs.}

\def\cvprPaperID{10490} % *** Enter the CVPR Paper ID here
\def\confName{CVPR}
\def\confYear{2022}

\begin{document}

\title{Marginal Contrastive Correspondence for Guided Image Generation}
% Coordinated

\author{
Fangneng Zhan\textsuperscript{\rm 1}  \
Yingchen Yu\textsuperscript{\rm 2}  \
Rongliang Wu\textsuperscript{\rm 2} \
Jiahui Zhang\textsuperscript{\rm 2}  \
Shijian Lu\thanks{Corresponding author} \ \textsuperscript{\rm 2} \
Changgong Zhang\textsuperscript{\rm 3}
\\
% \textsuperscript{\rm 1} 
\textsuperscript{\rm 1} S-Lab, Nanyang Technological University
\quad
\textsuperscript{\rm 2} Nanyang Technological University
\quad
\textsuperscript{\rm 3} Amazon
% Academy, Alibaba Group
% Nanyang Technological University   
}

\maketitle

\begin{abstract}
Exemplar-based image translation establishes dense correspondences between a conditional input and an exemplar (from two different domains) for leveraging detailed exemplar styles to achieve realistic image translation. Existing work builds the cross-domain correspondences implicitly by minimizing feature-wise distances across the two domains. Without explicit exploitation of domain-invariant features, this approach may not reduce the domain gap effectively which often leads to sub-optimal correspondences and image translation. We design a Marginal Contrastive Learning Network (MCL-Net) that explores contrastive learning to learn domain-invariant features for realistic exemplar-based image translation. Specifically, we design an innovative marginal contrastive loss that guides to establish dense correspondences explicitly. Nevertheless, building correspondence with domain-invariant semantics alone may impair the texture patterns and lead to degraded texture generation. We thus design a Self-Correlation Map (SCM) that incorporates scene structures as auxiliary information which improves the built correspondences substantially. Quantitative and qualitative experiments on multifarious image translation tasks show that the proposed method outperforms the state-of-the-art consistently.
\end{abstract}

\section{Introduction}
Image-to-image translation refers to the image generation conditioned on the certain inputs from other domains \cite{park2019spade,wang2018pix2pixhd,tang2019cycle}, and has delivered impressive advances with the emergence of Generative Adversarial Networks (GANs) \cite{goodfellow2014gan} in recent years. As a typical ill-posed problem, diverse solutions are naturally allowed in image translation tasks as one conditional input could corresponds to multiple image instances. Faithful control of the generation style not only enables diverse generation given certain conditions but also flexible user control of the desired generation. However, yielding high-fidelity images with controllable styles still remains a grand challenge. 
% -yingchen

\begin{figure}[t]
\centering
\includegraphics[width=1.0\linewidth]{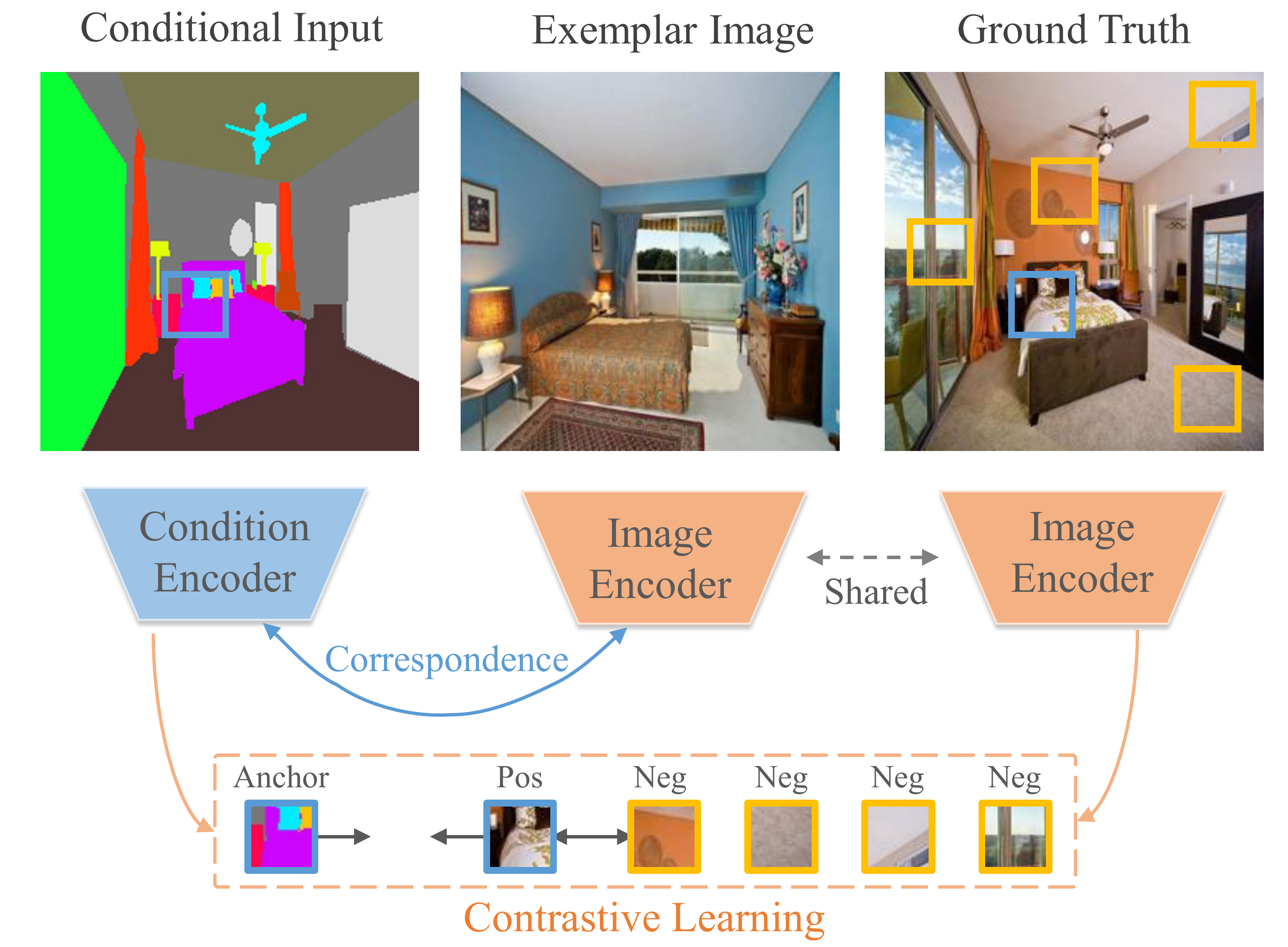}
\caption{
Learning domain-invariant features for building correspondence across domains: We exploit contrastive learning between the \textit{Conditional Input} and the \textit{Ground Truth} by pulling features at the same position closer while pushing those at different positions apart. With the learned \textit{Condition Encoder} and \textit{Image Encoder}, explicit feature correspondences can be built up between the \textit{Conditional Input} and the \textit{Exemplar Image}.
}
\label{im_intro}
\end{figure}

To tackle the style control challenge, one predominant approach employs variational autoencoders (VAEs) to regularize the image style into a Gaussian distribution as in SPADE \cite{park2019spade}, yet this approach often suffer from compromised generation quality due to the \textit{posterior collapse} phenomenon \cite{lucas2019don} of VAEs. Another direction is to leverage the styles extracted from the exemplars for guiding the generation. For instance, Zhu \emph{et al.} \cite{zhu2020sean} propose to achieve style control for each semantic region with a dedicated style code. However, the latent codes are typically extracted via a style encoder, which often reflects the global styles only and struggles to capture the detailed structures.
Very recently, dense correspondence has been actively explored in exemplar-based image translation, and emerges as a promising approach to achieve faithful style control. Specifically, Zhang \emph{et al.} \cite{zhang2020cocosnet} propose to establish dense semantic correspondence between conditioned input and a given exemplar so as to offer dense style guidance in translation. Zheng \emph{et al.} \cite{zheng2020semantic} and CoCosNet v2 \cite{zhou2021cocosnetv2} enables the establishment of dense correspondence in high-resolution features via PatchMatch to further enhance the preservation of detailed styles. 
Zhan \emph{et al.} \cite{zhan2021unite} introduce a general image translation framework that incorporates optimal transport for feature alignment between conditional inputs and style exemplars.
As a matching problem across very different domains, the key of building their correspondence lies in how to effectively learn the domain-invariant features to facilitate the proper matching. However, the aforementioned methods implicitly align the domain gaps for correspondence building by minimizing a pseudo pair loss and a L1 loss between features of condition and real image, which does not explicitly model the domain invariant features and may lead to sub-optimal feature correspondence.

In this paper, we propose \textbf{M}arginal \textbf{C}ontrastive \textbf{L}earning \textbf{Net}work (\textbf{MCL-Net}) for exemplar-based image translation, which introduce contrastive learning \cite{chen2020simple,tian2019contrastive,he2020momentum} to effectively extract domain-invariant features for building cross-domain correspondence as shown in Fig. \ref{im_intro}. In particular, contrastive learning is applied to the features of the conditional input and its ground truth that are extracted by separate encoders. Each input feature vector is treated as an \textit{anchor}, while the feature vector in the same spatial location is treated as \textit{positive sample} and the remaining feature vectors as \textit{negative samples}. By maximizing the mutual information between condition features and image features, contrastive learning could explicitly yield the domain-invariant features. Furthermore, a marginal contrastive loss (MCL) is proposed to enhance the discriminability of domain-invariant features, which could effectively curb over-smoothed or inaccurate correspondence. A deviation angle in MCL serves as a penalty to the positive anchor for margin preserving.

On the other hand, most previous methods build the correspondence relying on the corresponding local features without knowing the scene structures.
It means the scene structure could be impaired while building correspondence, which leads to distorted texture patterns and degraded generation performance.
Therefore, the scene structures such as the object shapes should be leveraged to facilitate the correspondence building, especially the preservation of fine texture patterns. Inspired by the observation that the self-attention maps could encode intact image structures even with appearance variations \cite{zheng2021spatially}, we design a self-correlation map (SCM) mechanism to explicitly represent the scene structures associated with feature and effectively facilitate the correspondence building.

The contributions of this work can be summarized in three aspects.
First, we introduce contrastive learning into the exemplar-based image translation framework to explicitly learn domain-invariant features for building correspondence.
Second, we propose a novel marginal contrastive loss to boost the feature discriminability in the representation space, which greatly benefits the establishment of explicit and accurate correspondence.
Third, we design a self-correlation map to properly represent the scene structures and effectively facilitate the preservation of texture patterns while building correspondence.

\section{Related Work}

\subsection{Image-to-Image Translation}

Due to the superior generation capability, GAN-based image-to-image translation \cite{zhu2017cyclegan,park2019spade,zhan2021unite,yu2021diverse,yu2021wavefill,zhan2019gadan,zhan2019sfgan,zhan2020aicnet} has been extensively investigated and achieved remarkable progress on translating different conditions such as semantic segmentation \cite{isola2017pix2pix,wang2018pix2pixhd,park2019spade,zhan2022modulated,zhan2021multimodal}, key points \cite{ma2017pose,men2020adgan,zhan2021emlight,zhan2021gmlight} and edge maps \cite{zhu2017toward,lee2018diverse,zhan2021rabit}.

As an ill-posed problem, image translation allows diverse contents or styles generation as long as they comply with the input conditions. However, the styles of generated images are typically determined by the learnt priors of large-scale datasets. To flexibly manipulate the styles for the diverse yet realistic image generation, optimal style control has attracted increasing attention recently. For instance, Huang \emph{et al.} \cite{huang2018multimodal} and Ma \emph{et al.} \cite{ma2018exemplar} extract and leverage style codes form the exemplar images to guide the generation process via adaptive instance normalization (AdaIN) \cite{huang2017adain}. Park \emph{et al.} \cite{park2019spade} offer a solution that unitizes variational autoencoder (VAE) \cite{kingma2013vae} to encode exemplars for image translation. A style encoder is introduced in Choi \emph{et al.} \cite{choi2020starganv2} to ensure the style consistency between the exemplars and the translated images. 
To specifically control the style in each semantic region, \cite{zhu2020sean} propose a semantic region-adaptive normalization (SEAN). 
Instead of leveraging latent codes for global style control, Zhang \emph{et al.} \cite{zhang2020cocosnet} propose a framework to enable finer style control by learning dense correspondence between conditional inputs and exemplars. On top of it, Zhang \emph{et al.} \cite{zhou2021cocosnetv2} introduce PatchMatch to facilitate the learning of dense correspondence on higher resolution, which further preserve better details or styles from the exemplars. However, most existing exemplar-based image translation methods implicitly learn the correspondence by directly apply feature-wised L1 loss to align the domain gaps. This strategy may lead to sub-optimal correspondence as the domain-invariant features are not explicitly explored. We propose to leverage the contrastive learning to explicitly learn domain-invariant features, and boost the discriminability of features via a novel marginal contrastive loss for building more explicit and accurate correspondence.

\begin{figure*}[t]
\centering
\includegraphics[width=1.0\linewidth]{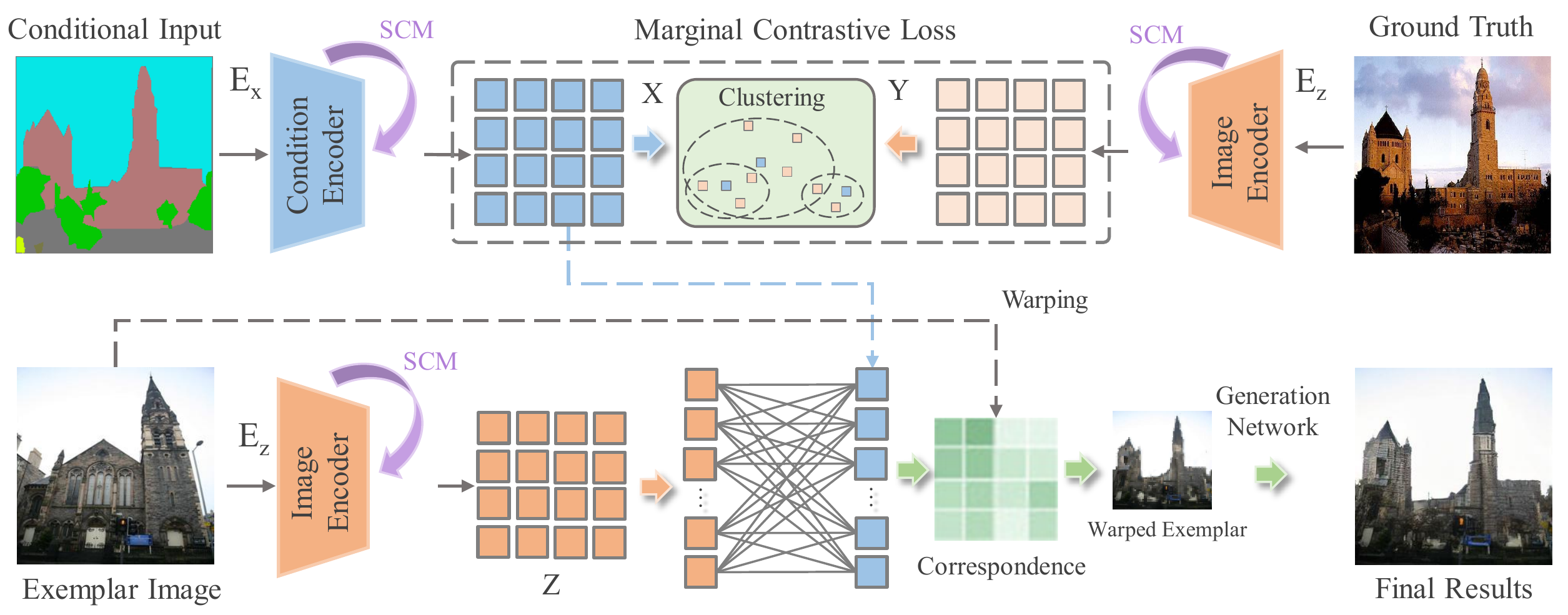}
\caption{
The framework of our proposed conditional image translation network: The \textit{Conditional Input} and \textit{Ground Truth} are fed to feature encoders $E_{X}$ and $E_{Z}$ to extract feature vectors $X$ and $Y$. The proposed self-correlation map (SCM) then encodes structure information for building correspondence, where the proposed \textit{Marginal Contrastive Loss} drives the encoders to learn domain-invariant features. With the shared feature encoder $E_{Z}$, domain-invariant features can be extracted from the \textit{Exemplar Image} and feature correspondences between the conditional input and exemplar can be established. The exemplar image can then be warped to align with the conditional input, which provides accurate style guidance for the \textit{Generation Network}. }
\label{im_stru}
\end{figure*}

\subsection{Contrastive Learning}

Recently, contrastive learning has shown its effectiveness in various computer vision tasks \cite{park2020contrastive,zhang2021blind}, especially in unsupervised representation learning \cite{sermanet2018time,tian2019contrastive,chen2020simple,he2020momentum,oord2018representation}. Its main idea is to learn a representation by pulling positive samples close to the anchor and pushing negative samples away. Different sampling strategies and contrastive losses have been extensively explored in various downstream talks. For example, Chen \emph{et al.} \cite{chen2020simple} and He \emph{et al.} \cite{he2020momentum} obtain the positive samples by augmenting the original data. Tian \emph{et al.} \cite{tian2019contrastive} treat multiple views of the same sample as positive pairs. On top of InfoNCE \cite{oord2018representation}, Park \emph{et al.} \cite{park2020contrastive} introduce PatchNCE to adopt contrastive learning in unpaired image-to-image translation, which treat image patches as samples.
In this work, we design a novel marginal contrastive loss based on InfoNCE, which could effectively enhance the discriminability of learned domain-invariant features to help build explicit and accurate correspondence.

\begin{figure*}[t]
\centering
\includegraphics[width=1.0\linewidth]{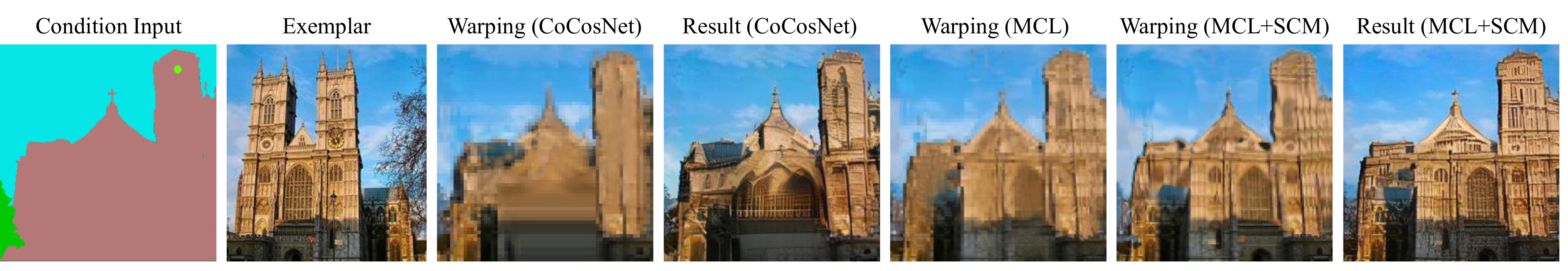}
\caption{
Comparison of warped exemplars and translation results by different methods: Warping and Result denote the warped exemplars and the final translation results. MCL and SCM denote our proposed marginal contrastive learning and self-correlation map, respectively.
}
\label{im_warp}
\end{figure*}

\section{Proposed Method}

Our MCL-Net consists of a correspondence network and a generation network as shown in Fig. \ref{im_stru}.
The correspondence network employs the proposed marginal contrastive learning to extract domain-invariant features for building correspondence, which is further utilized to warp the exemplars to be aligned with conditional inputs. 
Under guidance of the warped exemplar and the conditional input, the generation network produces the final translation results. More details to be described in the following subsections.

\subsection{Correspondence Network}

The correspondence network employs two encoders $E_X$ and $E_Z$ to encode the features of conditional inputs and exemplar images \& ground truth, respectively.
With the encoded features $X=[x_1, x_2, \cdots, c_N]$ and $Z=[z_1, z_2, \cdots, z_N]$ of the conditional input and exemplar image, the correspondence between them can be established by computing the feature-wise Cosine similarity.

To build the feature correspondence effectively between cross-domain images, domain-invariant features is expected to be extracted by the encoders.
Previous methods \cite{zhang2020cocosnet,zhou2021cocosnetv2} mainly employ a pseudo pairs loss and a L1 loss between conditional inputs and exemplars to drive the learning of feature encoders.
However, the pseudo pairs loss does not explicitly encourage the learning of domain-invariant features, while the L1 loss simply minimizes the distance between features in the same spatial location and ignores the distance between features in different spatial locations.
We thus introduce contrastive learning to learn domain-invariant features explicitly for building correspondence.

\subsection{Contrastive Correspondence}

Contrastive learning serves as a powerful tool for unsupervised representation learning by pulling positive samples closer and pushing negative samples apart.
To construct the positive and negative pairs for contrastive learning, we feed the ground truth image into encoder $E_Z$ to obtain encoded features $Y=[y_1, y_2, \cdots, y_N]$.
With a feature $x_i$ from the conditional input as the anchor, the corresponding feature $y_i$ from the ground truth serves as the positive samples, and the remaining $N-1$ features from the ground truth serve as the negative samples.
We normalize vectors onto a unit sphere to prevent the space from collapsing or expanding.
Then a noise contrastive estimation \cite{oord2018representation} framework is employed to conduct contrastive learning between the conditional input and ground truth as below:
\begin{equation}
\label{patchnce}
    \mathcal{L}_{xy} = - \sum_{i=1}^{N} \log \frac{ \exp( \frac{x_i \cdot y_i}{\tau} )  }{ \exp( \frac{x_i \cdot y_i}{\tau} ) + \sum_{\substack{j=1 \\ j\neq i}}^{N} \exp( \frac{x_i \cdot y_j}{\tau} ) }
\end{equation}
where $\tau$ denotes the temperature parameter, `.' stands for dot product of vectors.
Then a bidirectional contrastive loss $\mathcal{L}_{cl}$ can be formulated by: $\mathcal{L}_{cl} = \mathcal{L}_{xy} + \mathcal{L}_{yx}$.

\begin{figure}[t]
\centering
\includegraphics[width=1.0\linewidth]{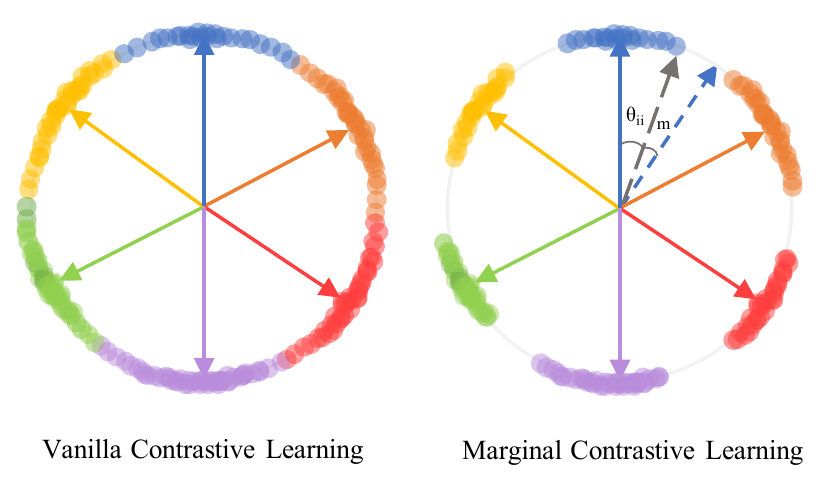}
\caption{
Toy illustration of learned feature representations with vanilla contrastive learning and our proposed marginal contrastive learning.
Dots indicate image features, angles $\theta_{ii}$ and $m$ denote the compactness of the learnt feature clusters and the angular margin penalty, respectively. Zoom-in for details.
}
\label{im_margin}
\end{figure}

\begin{algorithm}[!htb]
\begin{algorithmic}
\STATE \textbf{Input:} Conditional input features X, ground truth features Y.
\begin{enumerate} \setlength{\itemsep}{-\itemsep}
\item $[x_1, x_2, \cdots, x_N] $ = L2Norm(X)
\item $[y_1, y_2, \cdots, y_N] $ = L2Norm(Y)
\item  $\cos{\theta_{ii}}$ = $x_{i} \cdot z_i$ 
\item $\theta_{ii}$ = $\arccos{ ( \cos{\theta_{ii}} ) } $
\item add margin: $\cos (\theta_{ii} + m )$
\item marginal contrastive loss (MCL): \\
$ - \sum_{i=1}^{N} \log [ \frac{ \exp(  \cos(\theta_{ii} + m) ) }
{ \exp( \cos ( \theta_{ii} + m ) ) + \sum_{\substack{j=1 \\ j\neq i}}^{N}  \exp( \cos{\theta_{ij}} ) } ] $
\end{enumerate}
\STATE \textbf{Output:} MCL.
\end{algorithmic}
\caption{The Pseudo-code of the proposed marginal contrastive loss (MCL).}
\label{mcl}
\end{algorithm}

\paragraph{Marginal Contrastive Learning}
The contrastive learning enables to encode the domain-invariant features, while building accurate correspondence between two sets of features requires a high discriminability between features.
The original contrastive learning tends to produce smooth transition between different feature clusters as given in \ref{mcl}, which may incur smoothed and inaccurate correspondence.
Inspired by the additive angular margin loss in ArcFace \cite{deng2019arcface}, we propose a marginal contrastive loss to enlarge the separability of features on a hyper-sphere, which yields more explicit and accurate correspondence as shown in Fig. \ref{im_warp}.

Specially, as the features for building correspondence are normalized, the loss term $\mathcal{L}_{xy}$ in Eq. (\ref{patchnce}) can be rewritten as:
\begin{align*}
    \mathcal{L}_{xy} = 
      - \sum_{i=1}^{N} \log \frac{ \exp(s \cos{\theta_{ii}} )  }{ \exp( s \cos{ \theta_{ii} } ) + \sum_{\substack{j=1 \\ j\neq i}}^{N} \exp( s \cos{ \theta_{ij} } ) } 
\end{align*}
Thus, the embedded features are distributed around each feature center on the hyper-sphere with a radius of $s$.
To simultaneously enhance the intra-class compactness and inter-class discrepancy, we add an additive angular margin penalty $m$ (m=0.4 by default) to the positive samples to form marginal contrastive loss (MCL) as below:
\begin{align*}
    - \sum_{i=1}^{N} \log \frac{ \exp(s \cos(\theta_{ii} + m) )  }{ \exp( s \cos( \theta_{ii} + m ) ) + \sum_{\substack{j=1 \\ j\neq i}}^{N} \exp( s \cos{ \theta_{ij} } ) } 
\end{align*} 
The proposed MCL impose angular margin penalty in the normalized hyper-sphere which effectively enlarges the feature separability as illustrated in Fig. \ref{im_margin}.

\begin{figure}[t]
\centering
\includegraphics[width=1.0\linewidth]{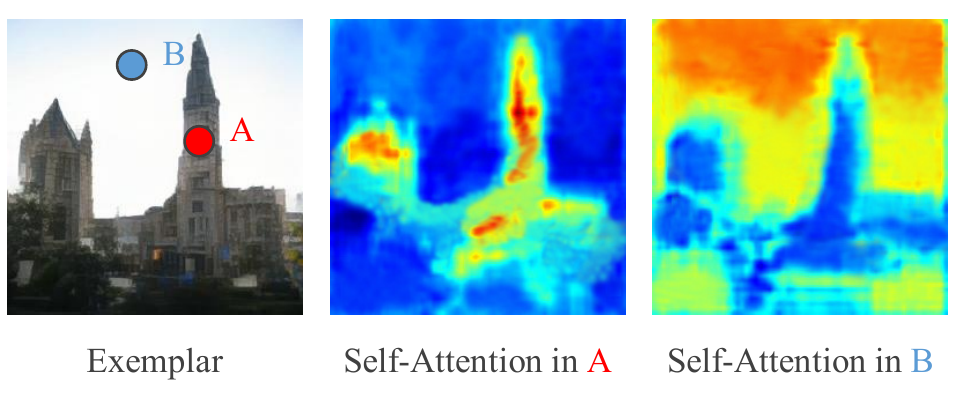}
\caption{
Illustration of the learnt self-correlation maps at two different spatial locations of the exemplar image.
}
\label{im_atten}
\end{figure}

\subsection{Self-Correlation Map}

Previous works \cite{zhang2020cocosnet,zhou2021cocosnetv2} build the correspondence relying on the corresponding local features without knowing the scene structure.
However, building correspondence purely relying on local features may impair the texture patterns of images, thus leading to degraded generation performance.
Intuitively, besides the local feature, scene structures (e.g., object shapes) can serve as auxiliary information to preserve the texture patterns and facilitate the building of correspondence.
We thus introduce a \textbf{S}elf-\textbf{C}orrelation \textbf{M}ap (\textbf{SCM}) in the correspondence network to utilize the scene structures for building correspondence.

As all regions within same categories exhibit some form of self-correlation, the self-correlation map of a feature effectively encodes the intact object shape information as shown in Fig. \ref{im_atten}.
By estimating these co-occurrence features for each spatial location, the scene structure of all locations can be explicitly represented.
Thus, given encoded features $X=[x_1, x_2, \cdots, x_N]$, the self-correlation map (SCM) of a feature can be denoted by:
\begin{equation}
    {\rm SCM_{i}} = (x_i)^T \cdot X, \quad i = 0,1,\cdots, N .
\end{equation}
Fig. \ref{im_atten} illustrates the sample of self-correlation maps.
As shown in the figure, the self-correlation maps accurately encode the scene structures in the corresponding location, e.g., the building and sky.

Unlike the original features that would encode various attributes such as style and texture, the self-correlation map only captures the spatial correlation.
To leverage SCM for building correspondence, the SCM ($64\times 64$ by default) is flatten into column vector (e.g., $4096$), followed by fully-connected layer to reduce the vector dimension to 256.
Then the feature vector of 256 dimension is concatenated with the encoded feature in corresponding spatial location to compute Cosine similarity for building correspondence.
As the proposed marginal contrastive learning is employed to learn domain-invariant feature for building correspondence, the learning of self-correlation maps is also driven by the designed contrastive loss.

\subsection{Generation Network}

With the warped exemplar to provide style guidance and the conditional input to provide semantic guidance, the generation network aims to generate high-fidelity images with faithful style to the exemplar and faithful semantic to the conditional input.
Specially, the warped exemplar is concatenated with the conditional input to be injected into the generation network through a spatially adaptive de-normalization \cite{park2019spade}.
The detailed architecture of the generation network is consistent with SPADE \cite{park2019spade}. More details of the network structure are available in the supplementary material.

\renewcommand\arraystretch{1.0}
\begin{table*}[t]
\renewcommand\tabcolsep{8pt}
\centering 
\begin{tabular}{l|ccc|ccc|ccc} \hline
\multirow{2}{*}{\textbf{Methods}}
& 
\multicolumn{3}{c|}{\textbf{ADE20K} \cite{zhou2017ade20k}} & 
\multicolumn{3}{c|}{\textbf{DeepFashion} \cite{liu2016deepfashion} } &
\multicolumn{3}{c}{\textbf{CelebA-HQ} \cite{liu2015celebahq}}
\\
\cline{2-10}
& FID $\downarrow$ & SWD $\downarrow$ & LPIPS $\uparrow$ 
& FID  $\downarrow$ & SWD  $\downarrow$ & LPIPS  $\uparrow$ 
& FID  $\downarrow$ & SWD  $\downarrow$ & LPIPS  $\uparrow$  \\\hline

\textbf{Pix2pixHD} \cite{wang2018pix2pixhd} & 81.80 & 35.70 & N/A      & 25.20  & 16.40  & N/A       & 42.70 & 33.30 & N/A     \\

\textbf{SPADE} \cite{park2019spade} & 33.90 & 19.70 & 0.344      & 36.20  & 27.80  & 0.231  & 31.50 & 26.90 & 0.187       \\

\textbf{SelectionGAN} \cite{tang2019selectiongan}    &  35.10 & 21.82 & 0.382          & 38.31  & 28.21  & 0.223     & 34.67 & 27.34 &  0.191   \\

\textbf{SMIS} \cite{zhu2020smis} & 42.17 &  22.67 &  0.416          & 22.23  & 23.73  & 0.240       &  23.71  &  22.23  &  0.201   \\

\textbf{SEAN}  \cite{zhu2020sean}   & 24.84 &  10.42  & 0.499       & 16.28  &  17.52  & 0.251       &  18.88    & 19.94 & 0.203      \\

\textbf{UNITE} \cite{zhan2021unite}
& 25.15 & 10.13 & \textbf{0.571}
& 13.08 & 16.65 & 0.278
& 13.15 & 14.91 & 0.213
  \\

\textbf{CoCosNet} \cite{zhang2020cocosnet}    & 26.40 & 10.50 & 0.560         & 14.40  & 17.20 & 0.272        & 14.30 & 15.30 & 0.208      \\
\textbf{CoCosNet v2} \cite{zhou2021cocosnetv2}    & 25.21 & 9.940 & 0.564         & \textbf{12.81}  & 16.53 & 0.283        & 12.85 & 14.62 & \textbf{0.218}      \\

\rowcolor{mygray} \textbf{MCL-Net}
& \textbf{24.75} & \textbf{9.852} & 0.569
& 12.89 & \textbf{16.24} & \textbf{0.286}
& \textbf{12.52} & \textbf{14.21} & 0.216
  \\
  \hline
\end{tabular}
\caption{
Comparing MCL-Net with state-of-the-art image translation methods: The comparisons were performed over three public datasets with three widely used evaluation metrics FID, SWD and LPIPS.
}
\label{tab_com}
\end{table*}

\renewcommand\arraystretch{1.25}
\begin{table}[t]
\small  
\renewcommand\tabcolsep{8pt}
\centering 
\begin{tabular}{
l|cc|cc
} \hline
\multirow{2}{*}{\textbf{Methods}} & 
\multicolumn{2}{c|}
{
Style Relevance
} & 
\multicolumn{2}{c}
{
Semantic
}  
\\
\cline{2-3}
 & Color & Texture &  \multicolumn{2}{c}
{
Consistency
}     \\\hline

\textbf{SPADE} \cite{park2019spade}  &  0.874   & 0.892  & \multicolumn{2}{c}{0.856}     \\

\textbf{SMIS} \cite{zhu2020smis} & 0.887     & 0.907 &  \multicolumn{2}{c}{0.858}     \\

\textbf{SEAN} \cite{zhu2020sean} &  0.932     & 0.926 &  \multicolumn{2}{c}{0.860}      \\ 

\textbf{UNITE} \cite{zhan2021unite} & 0.963 & 0.945 &  \multicolumn{2}{c}{0.869}     \\

\textbf{CoCosNet} \cite{zhang2020cocosnet} & 0.962 & 0.941 &  \multicolumn{2}{c}{0.862}      \\

\textbf{CoCosNet v2} \cite{zhou2021cocosnetv2} & \textbf{0.970} & 0.948 &  \multicolumn{2}{c}{0.877}     \\

% \hline
\rowcolor{mygray} \textbf{MCL-Net}  & 0.966 & \textbf{0.951} & \multicolumn{2}{c}{\textbf{0.881}}     \\\hline
\end{tabular}
\caption{
Quantitative evaluation of style relevance (color and texture) and semantic consistency on ADE20K.
}
\label{tab_consistency}
\end{table}

\subsection{Loss Functions}

The correspondence network and generation network are optimized jointly to learn the cross-domain correspondence and achieve high-fidelity image generation.
The conditional input, ground truth, exemplar are denoted by $X$, $Y$ and $Z$, respectively.
The condition encoder and image encoder in correspondence network are denoted by $E_{X}$ and $E_{Z}$, and the generator and discriminator are denoted by $G$ and $D$.

\paragraph{Correspondence Network}
Besides the proposed marginal contrastive loss $\mathcal{L}_{mcl}$, several other losses are designed to facilitate the learning of cross-domain correspondence.
As driven by the contrastive learning, the two feature encoder $E_{X}$ and $E_{Z}$ aim to extract domain invariant features (i.e., semantic features), the encoded features from the conditional input $X$ and the corresponding ground truth $Y$ should be consistent, thus yielding a feature consistency loss as defined below:
\begin{equation}
    \mathcal{L}_{fcst}= || E_{X}(X) - E_{Z}(Y) ||_{1}.
\end{equation}
To preserve image information in the warping process, the original exemplar should be recoverable from the warped exemplar through an inverse warping. Thus, a cycle-consistency loss can be formulated as below:
\begin{equation}
    \mathcal{L}_{cyc} = || T^{\top} \cdot T\cdot Z - Z ||_{1}
\end{equation}
where $T$ is the correspondence matrix.
Although it is hard to collect the ground truth of warped exemplars, pseudo exemplar pairs can be obtained by treating an augmented real image $Y'$ as the exemplar \cite{zhang2020cocosnet}.
We thus can penalize the difference between the warped exemplar and the augmented real image $Y'$ as below:
\begin{equation}
    \mathcal{L}_{pse} = || T\cdot Z - Y' ||_{1}
\end{equation}

\paragraph{Generation Network}
To achieve high-fidelity image translation, several losses are adopted in the generation network to work collaboratively.
The generated image $G(X,Z)$ should be consistent with the ground truth $Y$ in term of the semantics, thus a perceptual loss $\mathcal{L}_{perc}$ \cite{johnson2016perceptual} is employed to minimize their semantic discrepancy as below: 
\begin{equation}
    \mathcal{L}_{perc} = || \phi_{l}(G(X,Z)) - \phi_{l}(Y) ||_{1}
\end{equation}
where $\phi_{l}$ denotes $l$ layer of pre-trained VGG-19 \cite{simonyan2014vgg} model. 
On the other hand, the generated image $G(X,Z)$ should be consistent with the exemplar in terms of the style.
Thus, a contextual loss as described in \cite{mechrez2018contextual} is formulated as below:
\begin{equation}
    \mathcal{L}_{cxt} = - \log( \sum_{l} \mu_{i} CX_{ij} ( \phi_{l} (Z), \phi_{l} (Y) ) ) 
\end{equation}
where  $\mu_{m}$ balances the terms of different VGG layers.

To generate fine-grained image details, an adversarial loss $\mathcal{L}_{adv}$ is introduced with a discriminator $D$.
Thus, the overall objective function of the model is:
\begin{equation}
\begin{split}
    \mathcal{L} = & \mathop{\min}\limits_{E_{X},E_{Z},G} \mathop{\max}\limits_{D} (\lambda_1 \mathcal{L}_{cyc} + \lambda_2 \mathcal{L}_{fcst} + \lambda_3 \mathcal{L}_{mcl}  \\
    & \lambda_4 \mathcal{L}_{perc} + \lambda_5 \mathcal{L}_{cxt} + \lambda_6  \mathcal{L}_{pse} + \lambda_7 \mathcal{L}_{adv}) \\
\end{split}
\end{equation}
where $\lambda$ is weighting parameter.
Detailed ablation study of the loss terms can be found in the supplementary material.

\section{Experiments}

\subsection{Experimental Settings}
\label{setting}

\paragraph{Datasets.}
We conduct exemplar-based image translation on three different tasks.

\noindent
$\bullet$ ADE20k \cite{zhou2017ade20k} consists of 20,210 training images associated with a 150-class segmentation mask. 
The image translation is conducted with semantic segmentation as the input on this dataset.

\noindent
$\bullet$ CelebA-HQ \cite{liu2015celebahq} collects 30,000 high quality face images. 
The experiment is conducted with face edge maps as the conditional inputs.
Canny edge detector is employed to retrieve the connected face landmarks as face edges.

\noindent
$\bullet$ DeepFashion \cite{liu2016deepfashion} contains 52,712 high-quality persons in fashion clothe. 
The experiment is conducted with the human key-points as the conditional input.
OpenPose \cite{cao2017realtime} is employed to retrieve the human pose key-points.

\begin{figure*}[t]
\centering
\includegraphics[width=1.0\linewidth]{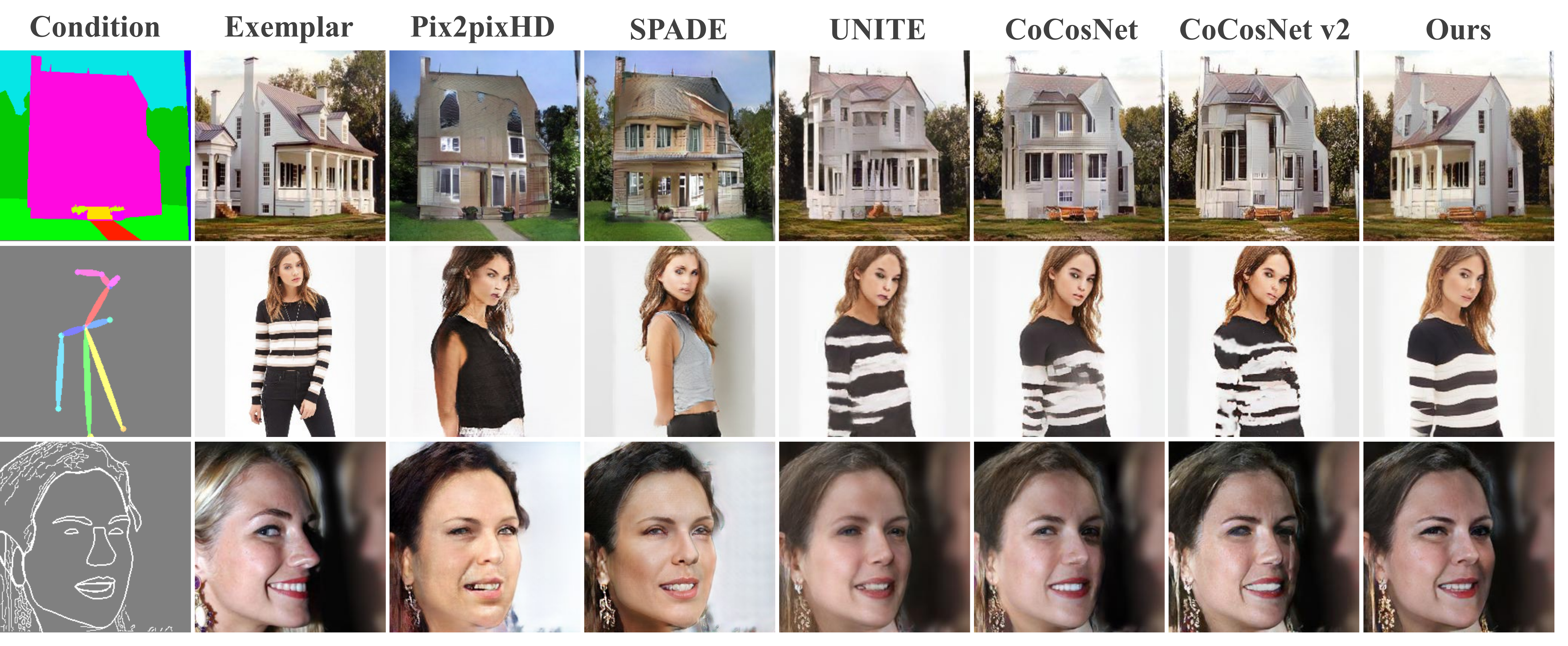}
\caption{
Qualitative comparison of MCL-Net and state-of-the-art image translation methods over three different types of conditional inputs including semantic segmentation, key-points, and edge map, respectively, from top to bottom.
}
\label{im_com}
\end{figure*}

\paragraph{Implementation Details.}

The correspondence network and generation network are jointly optimized with learning rates 1$e$-4 and 4$e$-4 for the generator and discriminator. 
Adam solver with $\beta_{1}=0$ and $\beta_{2}=0.999$ is adopted for the optimization. 
All experiments were conducted on 4 32GB Tesla V100 GPUs. 
The resolution for building correspondence is $64 \times 64$. 
The resolution of generated images is $256 \times 256$ for all translation tasks.

% \paragraph{Compared Methods.}
% We compare MCL-Net with several state-of-the-art translation methods including conditional image translation methods Pix2pixHD \cite{wang2018pix2pixhd}, SPADE \cite{park2019spade}, SelectionGAN \cite{tang2019selectiongan}, SMIS \cite{zhu2020smis}, SEAN \cite{zhu2020sean}, and exemplar-based image translation methods UNITE \cite{zhan2021unite}, CoCosNet \cite{zhang2020cocosnet}, and CoCosNet v2 \cite{zhou2021cocosnetv2}.

\subsection{Quantitative Evaluation}

\paragraph{Evaluation Metrics.} 
To evaluate the translation results comprehensively, 
\textit{Fr{\'e}chet Inception Score (FID)} \cite{fid} and \textit{Sliced Wasserstein distance (SWD)} \cite{swd} to evaluate the image perceptual quality;
\textit{Learned Perceptual Image Patch Similarity (LPIPS)} \cite{zhang2018lpips} is employed to evaluate the diversity of translated images.
Besides, L1 distance, peak signal-to-noise ratio (PSNR) and structural similarity index (SSIM) \cite{wang2004image} are also adopted to evaluate the low-level quality of warped images.

To evaluate the style relevance and semantic consistency of translated images, the metrics described in \cite{zhang2020cocosnet} is employed in this work.
Specially, as low-level features ($relu1\_2, relu2\_2$) usually encode image styles such as color and texture, we thus take the average cosine similarity of these layers in pre-trained VGG network \cite{simonyan2014very} as the score of style relevance ($relu1\_2$ for color, $relu2\_2$ for texture).
As high-level features ($relu3\_2, relu4\_2$, relu5\_2) tend to encode semantic features, we take the average cosine similarity of these layers in pre-trained VGG network \cite{simonyan2014very} as the score of semantic consistency.

\paragraph{Experimental Results.}
As shown in Table \ref{tab_com}, 
we can observe that MCL-Net outperforms nearly all compared methods in terms of image quality as measured by FID and SWD and image diversity as measured by LPIPS.
Compared with CoCosNet, MCL-Net achieves better FID and SWD as the designed marginal contrastive loss allows to learn domain invariant features for building correspondence and the self-correlation allows to incorporate structure information for building correspondence.

As an exemplar-based image translation task, the generated image should present semantic consistency with conditional inputs and present styles relevance with exemplars.
The evaluation results of semantic consistency and style consistence are tabulated in Table. \ref{tab_consistency}.
With our marginal contrastive learning for invariant feature learning and SCM for effective utilization of structure information, MCL-Net achieves the best feature correspondence which leads to best style relevance and semantic consistency.

\begin{figure}[t]
\centering
\includegraphics[width=1.0\linewidth]{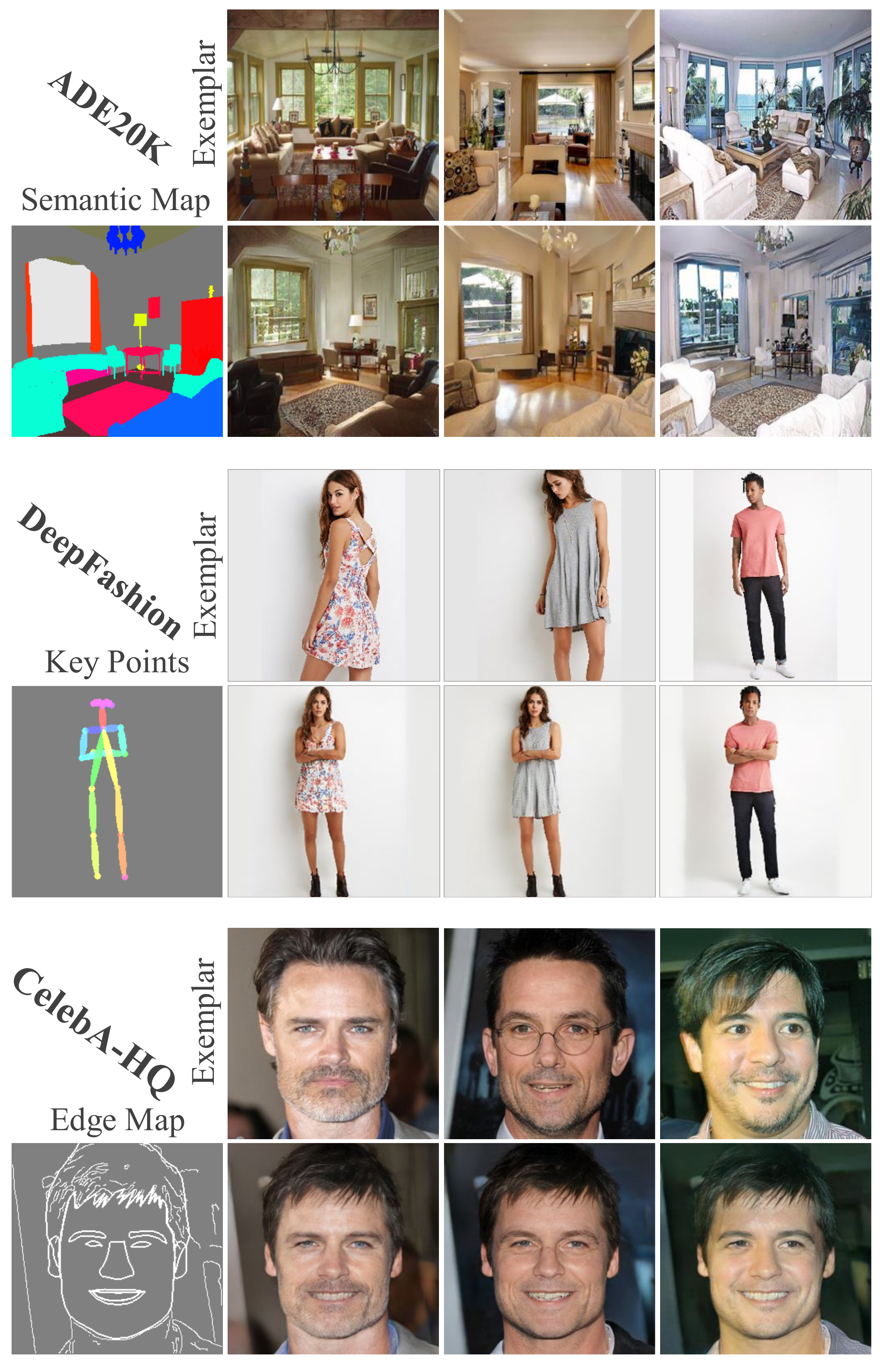}
\caption{
Qualitative illustration of our proposed MCL-Net with different types of conditional inputs and exemplars.
}
\label{im_diverse}
\end{figure}

\subsection{Qualitative Evaluation}

Fig. \ref{im_com} shows qualitative comparisons of different methods.
It can be observed that MCL-Net achieves the best visual quality with faithful styles as exemplars. 
Pix2pixHD \cite{wang2018pix2pixhd} is unable to capture the style from the exemplars.
SPADE \cite{park2019spade}, SMIS \cite{zhu2020smis} adopt variational auto-encoder (VAE) \cite{kingma2013auto} to encode image styles, while this method is unable to capture detailed style feature such as texture and patterns.
Exemplar-based translation methods CoCosNet \cite{zhang2020cocosnet} and CoCosNet v2 \cite{zhou2021cocosnetv2}
build cross-domain correspondence to capture detailed style from the exemplars, but they tend to produce blurred generation results without explicitly learning of domain invariant features and lose texture patterns without leveraging image structure information.
Our MCL-Net designs MCL to learn domain invariant features explicitly and introduce SCM to leverage the structure information for building correspondence.

Besides image quality, MCL-Net also demonstrates superior diversity in image translation as illustrated in Fig. \ref{im_diverse}.
We can observe that MCL-Net is capable of synthesizing realistic images with faithful style to the given exemplars.

\renewcommand\arraystretch{1.2}
\begin{table}[t]
\small
\renewcommand\tabcolsep{9pt}
\centering 
\begin{tabular}{
l|ccc
} 
\hline
\textbf{Models} & \textbf{FID}  $\downarrow$ & \textbf{SWD}  $\downarrow$ & \textbf{LPIPS}  $\uparrow$ 
\\
\cline{2-3}
\hline\hline
\textbf{Baseline}      & 32.02  &  26.87  &  0.184    \\
\textbf{+COR}      & 15.77  &  16.31  &  0.199    \\
\textbf{+COR+MCL}     & 12.75  & 15.03  &  0.208   \\
\textbf{+COR+SCM}      & 13.27  & 15.24  &  \textbf{0.219}   \\
\rowcolor{mygray} \textbf{+COR+MCL+SCM}     & \textbf{12.52}  & \textbf{14.21}  &  0.216   \\
\hline
\end{tabular}
\caption
{
Ablation studies of our MCL-Net designs over CelebA-HQ \cite{liu2015celebahq}: The baseline is SPADE \cite{park2019spade} that conduct translation without building correspondence. 
COR denotes building correspondence between conditional inputs and exemplars.
MCL and SCM mean to include the proposed marginal contrastive loss and self-correlation map in building correspondence. 
Model in the last row is the standard MCL-Net.
}
\label{tab_ablation}
\end{table}

\renewcommand\arraystretch{1.2}
\begin{table}[t]
\small
\renewcommand\tabcolsep{9pt}
\centering 
\begin{tabular}{
l|ccc
} 
\hline
\textbf{Models} & \textbf{L1}  $\downarrow$ & \textbf{PSNR}  $\uparrow$ & \textbf{SSIM}  $\uparrow$ 
\\
\cline{2-3}
\hline\hline
\textbf{Baseline}        & 80.14  & 28.35  &  0.766    \\
\textbf{+MCL (m=0.1)}      & 79.32  & 29.01  &  0.774   \\
\textbf{+MCL (m=0.2)}     & 78.05  & 29.44  &  0.787   \\
\textbf{+MCL (m=0.3)}     & 76.43  & 29.75  &  0.795   \\
\textbf{+MCL (m=0.4)}     & 75.27  & 30.04  &  0.811   \\
\textbf{+SCM}            & 76.07  & 29.67  &  0.791   \\
\rowcolor{mygray} \textbf{+MCL (m=0.4)+SCM}     & \textbf{73.72}  & \textbf{30.94}  &  \textbf{0.826}   \\
\hline
\end{tabular}
\caption
{
Parameter studies of marginal contrastive learning (MCL) and self-correlation map (SCM) on warped exemplars on DeepFashion dataset. The baseline is CoCosNet\cite{zhang2020cocosnet}, and the last row denotes the standard MCL-Net.
}
\label{tab_ablation2}
\end{table}

\subsection{Discussion}

We conduct extensive ablation studies to evaluate our technical designs.
Table \ref{tab_ablation} shows experimental results on CelebA-HQ.
SPADE \cite{park2019spade} is selected as the baseline which achieves image translation \& manipulation without feature alignment. The performance is clearly improved when feature correspondence (COR) is included to align features (namely CoCosNet \cite{zhang2020cocosnet}). 
The generation performance can be further improved by including the proposed marginal contrastive learning (MCL) to learn domain invariant features or the self-correlation map (SCM) for building correspondence.
Including both MCL and SCM to the model leads to the best generation performance.

To directly evaluate the accuracy of the built correspondence, we evaluate the warped exemplars on DeepFashion dataset where paired data are available by treating person images under different poses as the exemplar and the ground truth.
Hence, we can measure the distance between the warped exemplars and the ground truth with L1, PSNR and SSIM \cite{wang2004image}. 
Specially, the margin parameter $m$ plays a key role in MCL, we thus conduct experiments to investigate the effect of $m$. 
As shown in Table \ref{tab_ablation2}, the vanilla correspondence is selected as the baseline.
By varying $m$ from 0.1 to 0.4, we can observe that the correspondence accuracy is improved consistently with the increasing of margin $m$.
Although large margin $m$ contributes to the correspondence accuracy, we find the model tends to be unstable and even fail to converge with a large margin, e.g., $m=0.5$.
We thus select $m=0.4$ as the default setting in our MCL.
Besides, including the designed self-correlation map also improve correspondence accuracy substantially, and combine MCL ($m=0.4$) and SCM yields the best correspondence accuracy.

\section{Conclusions}
This paper presents MCL-Net, an exemplar-based image translation framework that introduces a marginal contrastive learning to learn domain invariant feature across conditional input and exemplar for building correspondence.
To encode the structure information to facilitate correspondence building, we propose a novel self-correlation map which captures co-occurrence features for each spatial location and represents the image structure explicitly.
Quantitative and qualitative experiments show that MCL-Net is capable of generating high-fidelity images with consistent semantic with the conditional input and faithful style to the exemplar.

\section{Acknowledgement}
This study is supported under the RIE2020 Industry Alignment Fund – Industry Collaboration Projects (IAF-ICP) Funding Initiative, as well as cash and in-kind contribution from the industry partner(s).

{\small
\bibliographystyle{ieee_fullname}
\bibliography{egbib}

\begin{thebibliography}{10}\itemsep=-1pt

\bibitem{cao2017realtime}
Zhe Cao, Tomas Simon, Shih-En Wei, and Yaser Sheikh.
\newblock Realtime multi-person 2d pose estimation using part affinity fields.
\newblock In {\em Proceedings of the IEEE conference on computer vision and
  pattern recognition}, pages 7291--7299, 2017.

\bibitem{chen2020simple}
Ting Chen, Simon Kornblith, Mohammad Norouzi, and Geoffrey Hinton.
\newblock A simple framework for contrastive learning of visual
  representations.
\newblock In {\em International conference on machine learning}, pages
  1597--1607. PMLR, 2020.

\bibitem{choi2020starganv2}
Yunjey Choi, Youngjung Uh, Jaejun Yoo, and Jung-Woo Ha.
\newblock Stargan v2: Diverse image synthesis for multiple domains.
\newblock In {\em Proceedings of the IEEE/CVF Conference on Computer Vision and
  Pattern Recognition}, pages 8188--8197, 2020.

\bibitem{deng2019arcface}
Jiankang Deng, Jia Guo, Niannan Xue, and Stefanos Zafeiriou.
\newblock Arcface: Additive angular margin loss for deep face recognition.
\newblock In {\em Proceedings of the IEEE/CVF Conference on Computer Vision and
  Pattern Recognition}, pages 4690--4699, 2019.

\bibitem{goodfellow2014gan}
Ian Goodfellow, Jean Pouget-Abadie, Mehdi Mirza, Bing Xu, David Warde-Farley,
  Sherjil Ozair, Aaron Courville, and Yoshua Bengio.
\newblock Generative adversarial nets.
\newblock In {\em Advances in neural information processing systems}, pages
  2672--2680, 2014.

\bibitem{he2020momentum}
Kaiming He, Haoqi Fan, Yuxin Wu, Saining Xie, and Ross Girshick.
\newblock Momentum contrast for unsupervised visual representation learning.
\newblock In {\em Proceedings of the IEEE/CVF Conference on Computer Vision and
  Pattern Recognition}, pages 9729--9738, 2020.

\bibitem{fid}
Martin Heusel, Hubert Ramsauer, Thomas Unterthiner, Bernhard Nessler, and Sepp
  Hochreiter.
\newblock Gans trained by a two time-scale update rule converge to a local nash
  equilibrium.
\newblock In {\em Advances in neural information processing systems}, pages
  6626--6637, 2017.

\bibitem{huang2017adain}
Xun Huang and Serge Belongie.
\newblock Arbitrary style transfer in real-time with adaptive instance
  normalization.
\newblock In {\em Proceedings of the IEEE International Conference on Computer
  Vision}, pages 1501--1510, 2017.

\bibitem{huang2018multimodal}
Xun Huang, Ming-Yu Liu, Serge Belongie, and Jan Kautz.
\newblock Multimodal unsupervised image-to-image translation.
\newblock In {\em Proceedings of the European Conference on Computer Vision
  (ECCV)}, pages 172--189, 2018.

\bibitem{isola2017pix2pix}
Phillip Isola, Jun-Yan Zhu, Tinghui Zhou, and Alexei~A Efros.
\newblock Image-to-image translation with conditional adversarial networks.
\newblock In {\em Proceedings of the IEEE conference on computer vision and
  pattern recognition}, pages 1125--1134, 2017.

\bibitem{johnson2016perceptual}
Justin Johnson, Alexandre Alahi, and Li Fei-Fei.
\newblock Perceptual losses for real-time style transfer and super-resolution.
\newblock In {\em European conference on computer vision}, pages 694--711.
  Springer, 2016.

\bibitem{swd}
Tero Karras, Timo Aila, Samuli Laine, and Jaakko Lehtinen.
\newblock Progressive growing of gans for improved quality, stability, and
  variation.
\newblock {\em arXiv preprint arXiv:1710.10196}, 2017.

\bibitem{kingma2013vae}
Diederik~P Kingma and Max Welling.
\newblock Auto-encoding variational bayes.
\newblock {\em arXiv preprint arXiv:1312.6114}, 2013.

\bibitem{kingma2013auto}
Diederik~P Kingma and Max Welling.
\newblock Auto-encoding variational bayes.
\newblock {\em arXiv preprint arXiv:1312.6114}, 2013.

\bibitem{lee2018diverse}
Hsin-Ying Lee, Hung-Yu Tseng, Jia-Bin Huang, Maneesh Singh, and Ming-Hsuan
  Yang.
\newblock Diverse image-to-image translation via disentangled representations.
\newblock In {\em Proceedings of the European conference on computer vision
  (ECCV)}, pages 35--51, 2018.

\bibitem{liu2016deepfashion}
Ziwei Liu, Ping Luo, Shi Qiu, Xiaogang Wang, and Xiaoou Tang.
\newblock Deepfashion: Powering robust clothes recognition and retrieval with
  rich annotations.
\newblock In {\em Proceedings of the IEEE conference on computer vision and
  pattern recognition}, pages 1096--1104, 2016.

\bibitem{liu2015celebahq}
Ziwei Liu, Ping Luo, Xiaogang Wang, and Xiaoou Tang.
\newblock Deep learning face attributes in the wild.
\newblock In {\em Proceedings of the IEEE international conference on computer
  vision}, pages 3730--3738, 2015.

\bibitem{lucas2019don}
James Lucas, George Tucker, Roger Grosse, and Mohammad Norouzi.
\newblock Don't blame the elbo! a linear vae perspective on posterior collapse.
\newblock {\em arXiv preprint arXiv:1911.02469}, 2019.

\bibitem{ma2018exemplar}
Liqian Ma, Xu Jia, Stamatios Georgoulis, Tinne Tuytelaars, and Luc Van~Gool.
\newblock Exemplar guided unsupervised image-to-image translation with semantic
  consistency.
\newblock In {\em International Conference on Learning Representations}, 2018.

\bibitem{ma2017pose}
Liqian Ma, Xu Jia, Qianru Sun, Bernt Schiele, Tinne Tuytelaars, and Luc
  Van~Gool.
\newblock Pose guided person image generation.
\newblock In {\em Advances in neural information processing systems}, pages
  406--416, 2017.

\bibitem{mechrez2018contextual}
Roey Mechrez, Itamar Talmi, and Lihi Zelnik-Manor.
\newblock The contextual loss for image transformation with non-aligned data.
\newblock In {\em Proceedings of the European Conference on Computer Vision
  (ECCV)}, pages 768--783, 2018.

\bibitem{men2020adgan}
Yifang Men, Yiming Mao, Yuning Jiang, Wei-Ying Ma, and Zhouhui Lian.
\newblock Controllable person image synthesis with attribute-decomposed gan.
\newblock In {\em Proceedings of the IEEE/CVF Conference on Computer Vision and
  Pattern Recognition}, pages 5084--5093, 2020.

\bibitem{oord2018representation}
Aaron van~den Oord, Yazhe Li, and Oriol Vinyals.
\newblock Representation learning with contrastive predictive coding.
\newblock {\em arXiv preprint arXiv:1807.03748}, 2018.

\bibitem{park2020contrastive}
Taesung Park, Alexei~A Efros, Richard Zhang, and Jun-Yan Zhu.
\newblock Contrastive learning for unpaired image-to-image translation.
\newblock In {\em European Conference on Computer Vision}, pages 319--345.
  Springer, 2020.

\bibitem{park2019spade}
Taesung Park, Ming-Yu Liu, Ting-Chun Wang, and Jun-Yan Zhu.
\newblock Semantic image synthesis with spatially-adaptive normalization.
\newblock In {\em Proceedings of the IEEE Conference on Computer Vision and
  Pattern Recognition}, pages 2337--2346, 2019.

\bibitem{sermanet2018time}
Pierre Sermanet, Corey Lynch, Yevgen Chebotar, Jasmine Hsu, Eric Jang, Stefan
  Schaal, Sergey Levine, and Google Brain.
\newblock Time-contrastive networks: Self-supervised learning from video.
\newblock In {\em 2018 IEEE International Conference on Robotics and Automation
  (ICRA)}, pages 1134--1141. IEEE, 2018.

\bibitem{simonyan2014vgg}
Karen Simonyan and Andrew Zisserman.
\newblock Very deep convolutional networks for large-scale image recognition.
\newblock {\em arXiv preprint arXiv:1409.1556}, 2014.

\bibitem{simonyan2014very}
Karen Simonyan and Andrew Zisserman.
\newblock Very deep convolutional networks for large-scale image recognition.
\newblock {\em arXiv preprint arXiv:1409.1556}, 2014.

\bibitem{tang2019cycle}
Hao Tang, Dan Xu, Gaowen Liu, Wei Wang, Nicu Sebe, and Yan Yan.
\newblock Cycle in cycle generative adversarial networks for keypoint-guided
  image generation.
\newblock In {\em Proceedings of the 27th ACM International Conference on
  Multimedia}, pages 2052--2060, 2019.

\bibitem{tang2019selectiongan}
Hao Tang, Dan Xu, Nicu Sebe, Yanzhi Wang, Jason~J Corso, and Yan Yan.
\newblock Multi-channel attention selection gan with cascaded semantic guidance
  for cross-view image translation.
\newblock In {\em Proceedings of the IEEE Conference on Computer Vision and
  Pattern Recognition}, pages 2417--2426, 2019.

\bibitem{tian2019contrastive}
Yonglong Tian, Dilip Krishnan, and Phillip Isola.
\newblock Contrastive multiview coding.
\newblock {\em arXiv preprint arXiv:1906.05849}, 2019.

\bibitem{wang2018pix2pixhd}
Ting-Chun Wang, Ming-Yu Liu, Jun-Yan Zhu, Andrew Tao, Jan Kautz, and Bryan
  Catanzaro.
\newblock High-resolution image synthesis and semantic manipulation with
  conditional gans.
\newblock In {\em Proceedings of the IEEE conference on computer vision and
  pattern recognition}, pages 8798--8807, 2018.

\bibitem{wang2004image}
Zhou Wang, Alan~C Bovik, Hamid~R Sheikh, and Eero~P Simoncelli.
\newblock Image quality assessment: from error visibility to structural
  similarity.
\newblock {\em IEEE transactions on image processing}, 13(4):600--612, 2004.

\bibitem{yu2021wavefill}
Yingchen Yu, Fangneng Zhan, Shijian Lu, Jianxiong Pan, Feiying Ma, Xuansong
  Xie, and Chunyan Miao.
\newblock Wavefill: A wavelet-based generation network for image inpainting.
\newblock In {\em Proceedings of the IEEE/CVF International Conference on
  Computer Vision}, pages 14114--14123, 2021.

\bibitem{yu2021diverse}
Yingchen Yu, Fangneng Zhan, Rongliang Wu, Jianxiong Pan, Kaiwen Cui, Shijian
  Lu, Feiying Ma, Xuansong Xie, and Chunyan Miao.
\newblock Diverse image inpainting with bidirectional and autoregressive
  transformers.
\newblock In {\em Proceedings of the 29th ACM International Conference on
  Multimedia}, pages 69--78, 2021.

\bibitem{zhan2020aicnet}
Fangneng Zhan, Shijian Lu, Changgong Zhang, Feiying Ma, and Xuansong Xie.
\newblock Adversarial image composition with auxiliary illumination.
\newblock In {\em Proceedings of the Asian Conference on Computer Vision},
  2020.

\bibitem{zhan2019gadan}
Fangneng Zhan, Chuhui Xue, and Shijian Lu.
\newblock Ga-dan: Geometry-aware domain adaptation network for scene text
  detection and recognition.
\newblock In {\em Proceedings of the IEEE International Conference on Computer
  Vision}, pages 9105--9115, 2019.

\bibitem{zhan2021unite}
Fangneng Zhan, Yingchen Yu, Kaiwen Cui, Gongjie Zhang, Shijian Lu, Jianxiong
  Pan, Changgong Zhang, Feiying Ma, Xuansong Xie, and Chunyan Miao.
\newblock Unbalanced feature transport for exemplar-based image translation.
\newblock In {\em Proceedings of the IEEE Conference on Computer Vision and
  Pattern Recognition}, 2021.

\bibitem{zhan2021rabit}
Fangneng Zhan, Yingchen Yu, Rongliang Wu, Kaiwen Cui, Aoran Xiao, Shijian Lu,
  and Ling Shao.
\newblock Bi-level feature alignment for semantic image translation \&
  manipulation.
\newblock {\em arXiv preprint}, 2021.

\bibitem{zhan2021gmlight}
Fangneng Zhan, Yingchen Yu, Rongliang Wu, Changgong Zhang, Shijian Lu, Ling
  Shao, Feiying Ma, and Xuansong Xie.
\newblock Gmlight: Lighting estimation via geometric distribution
  approximation.
\newblock {\em arXiv preprint arXiv:2102.10244}, 2021.

\bibitem{zhan2021multimodal}
Fangneng Zhan, Yingchen Yu, Rongliang Wu, Jiahui Zhang, and Shijian Lu.
\newblock Multimodal image synthesis and editing: A survey.
\newblock {\em arXiv preprint arXiv:2112.13592}, 2021.

\bibitem{zhan2021emlight}
Fangneng Zhan, Changgong Zhang, Yingchen Yu, Yuan Chang, Shijian Lu, Feiying
  Ma, and Xuansong Xie.
\newblock Emlight: Lighting estimation via spherical distribution
  approximation.
\newblock In {\em Proceedings of the AAAI Conference on Artificial
  Intelligence}, pages 3287--3295, 2021.

\bibitem{zhan2022modulated}
Fangneng Zhan, Jiahui Zhang, Yingchen Yu, Rongliang Wu, and Shijian Lu.
\newblock Modulated contrast for versatile image synthesis.
\newblock {\em arXiv preprint arXiv:2203.09333}, 2022.

\bibitem{zhan2019sfgan}
Fangneng Zhan, Hongyuan Zhu, and Shijian Lu.
\newblock Spatial fusion gan for image synthesis.
\newblock In {\em Proceedings of the IEEE conference on computer vision and
  pattern recognition}, pages 3653--3662, 2019.

\bibitem{zhang2021blind}
Jiahui Zhang, Shijian Lu, Fangneng Zhan, and Yingchen Yu.
\newblock Blind image super-resolution via contrastive representation learning.
\newblock {\em arXiv preprint arXiv:2107.00708}, 2021.

\bibitem{zhang2020cocosnet}
Pan Zhang, Bo Zhang, Dong Chen, Lu Yuan, and Fang Wen.
\newblock Cross-domain correspondence learning for exemplar-based image
  translation.
\newblock In {\em Proceedings of the IEEE/CVF Conference on Computer Vision and
  Pattern Recognition}, pages 5143--5153, 2020.

\bibitem{zhang2018lpips}
Richard Zhang, Phillip Isola, Alexei~A Efros, Eli Shechtman, and Oliver Wang.
\newblock The unreasonable effectiveness of deep features as a perceptual
  metric.
\newblock In {\em Proceedings of the IEEE conference on computer vision and
  pattern recognition}, pages 586--595, 2018.

\bibitem{zheng2021spatially}
Chuanxia Zheng, Tat-Jen Cham, and Jianfei Cai.
\newblock The spatially-correlative loss for various image translation tasks.
\newblock In {\em Proceedings of the IEEE/CVF Conference on Computer Vision and
  Pattern Recognition}, pages 16407--16417, 2021.

\bibitem{zheng2020semantic}
Haitian Zheng, Zhe Lin, Jingwan Lu, Scott Cohen, Jianming Zhang, Ning Xu, and
  Jiebo Luo.
\newblock Semantic layout manipulation with high-resolution sparse attention.
\newblock {\em arXiv preprint arXiv:2012.07288}, 2020.

\bibitem{zhou2017ade20k}
Bolei Zhou, Hang Zhao, Xavier Puig, Sanja Fidler, Adela Barriuso, and Antonio
  Torralba.
\newblock Scene parsing through ade20k dataset.
\newblock In {\em Proceedings of the IEEE conference on computer vision and
  pattern recognition}, pages 633--641, 2017.

\bibitem{zhou2021cocosnetv2}
Xingran Zhou, Bo Zhang, Ting Zhang, Pan Zhang, Jianmin Bao, Dong Chen, Zhongfei
  Zhang, and Fang Wen.
\newblock Cocosnet v2: Full-resolution correspondence learning for image
  translation.
\newblock In {\em Proceedings of the IEEE/CVF Conference on Computer Vision and
  Pattern Recognition}, pages 11465--11475, 2021.

\bibitem{zhu2017cyclegan}
Jun-Yan Zhu, Taesung Park, Phillip Isola, and Alexei~A Efros.
\newblock Unpaired image-to-image translation using cycle-consistent
  adversarial networks.
\newblock In {\em Proceedings of the IEEE international conference on computer
  vision}, pages 2223--2232, 2017.

\bibitem{zhu2017toward}
Jun-Yan Zhu, Richard Zhang, Deepak Pathak, Trevor Darrell, Alexei~A Efros,
  Oliver Wang, and Eli Shechtman.
\newblock Toward multimodal image-to-image translation.
\newblock In {\em Advances in neural information processing systems}, pages
  465--476, 2017.

\bibitem{zhu2020sean}
Peihao Zhu, Rameen Abdal, Yipeng Qin, and Peter Wonka.
\newblock Sean: Image synthesis with semantic region-adaptive normalization.
\newblock In {\em Proceedings of the IEEE/CVF Conference on Computer Vision and
  Pattern Recognition}, pages 5104--5113, 2020.

\bibitem{zhu2020smis}
Zhen Zhu, Zhiliang Xu, Ansheng You, and Xiang Bai.
\newblock Semantically multi-modal image synthesis.
\newblock In {\em Proceedings of the IEEE/CVF Conference on Computer Vision and
  Pattern Recognition}, pages 5467--5476, 2020.

\end{thebibliography}
}

\end{document}